# Using MRNet to Predict Lunar Rock Categories Detected by Chang'e 5 Probe

Jin CUI [1]

*School of Information and Communication Engineering, Beijing University of Posts and Telecommunications, Beijing, China*

Yifei ZOU

*School of Environment, China University of Geosciences, Wuhan, China*

Siyuan ZHANG

*Faculty of Engineering, The University of Sydney, Sydney, Australia*

**Abstract.** China's Chang'e 5 mission has been a remarkable success, with the chang'e 5 lander traveling on the Oceanus Procellarum to collect images of the lunar surface. Over the past half century, people have brought back some lunar rock samples, but its quantity does not meet the need for research. Under current circumstances, people still mainly rely on the analysis of rocks on the lunar surface through the detection of lunar rover. The Oceanus Procellarum, chosen by Chang'e 5 mission, contains various kind of rock species. Therefore, we first applied to the National Astronomical Observatories of the China under the Chinese Academy of Sciences for the Navigation and Terrain Camera (NaTeCam) of the lunar surface image, and established a lunar surface rock image data set CE5ROCK. The data set contains 100 images, which randomly divided into training, validation and test set. Experimental results show that the identification accuracy testing on convolutional neural network (CNN) models like AlexNet or MobileNet is about to 40.0%. In order to make full use of the global information in Moon images, this paper proposes the MRNet (MoonRockNet) network architecture. The encoding structure of the network uses VGG16 for feature extraction, and the decoding part adds dilated convolution and commonly used U-Net structure on the original VGG16 decoding structure, which is more conducive to identify more refined but more sparsely distributed types of lunar rocks. We have conducted extensive experiments on the established CE5ROCK data set, and the experimental results show that MRNet can achieve more accurate rock type identification, and outperform other existing mainstream algorithms in the identification performance.

## 1. Introduction

In recent years, the moon has been one of the most popular missions in space exploration. National Aeronautics and Space Administration (NASA) has launched several lunar landing missions, and Apollo missions carried out lunar surface exploration missions [1].China's first lunar sampling return mission is the Chang' e 5 mission which consists of [2]: orbiting, landing and returning missions. The moon rover has successfully conducted a Moon science exploration mission on the lunar surface for more than a year, and sent back the multi-dimensional lunar science data [3]. The deep learning-based rock type identification algorithm proposed in this paper provides an idea to solve the above problems.

---
[1] Jin Cui, Corresponding author, School of Information and Communication Engineering, Beijing University of Posts and Telecommunications, Beijing, China; E-mail: 15611298178@163.com

Furthermore, there are few publicly published deep learning models [4] specifically for surface topography recognition of extraterrestrial rocks.The main contributions of this work can be summarized as follows:

1. Presenting a MRNet suitable for the classification of lunar rock types, based on existing deep learning algorithms. This network fused U-Net architecture to traditional convolutional neural networks.

2. A hybrid dilated convolution is also introduced in the decoding part of the network, which performs well at high input image resolution. In addition, this paper establishes a lunar surface image classification data set CE5ROCK based on the lunar surface images collected by the Chang'e 5 lander lunar rover navigation terrain camera.

3. The data set was manually screened and annotated, and consists of 160 images of lunar rock. There is an Earth rock data set for comparison.

The rest of paper is organized as follows. Firstly, in Sect.2 we showed explicit methods in image recognition tasks, including some classic deep learning networks, and gave detailed analysis and explanation of U-Net and VGG16. In Sect.3 we introduced the lunar surface rock image data set and elucidated the results in tables and charts, which show the analyzing outcome explicitly. The performance of the MRNet network was analyzed using f-score and ROC curve. Finally, Sect.4 concludes this paper.

**2. Methodology**

2.1. Moon Rock Type

In previous studies, some scholars have divided lunar lithology into igneous rocks according to sample rock types, crystalline impact lava, impact glass, hot metamorphic rocks and polyclastic matrix horn rocks (Heiken et.al, 1991[5]). Some scholars have divided the rock into minerals formed by silicate magma and molten conglomerate formed under the impact of meteorites. Newer views divide lunar rocks into lunar sea basalt, highland plagiosic, KREEP, and other possible rocks. Based on the sample image shown by NASA, which are divided into basalt and breccia, and initially classified to play a screening role for subsequent studies.

2.2 MRNet(MoonRockNet)

This section presents a Moon Rock specified classification network that combines U-Net with the VGG16 framework. The VGG16 encoding layer was placed after the ResNet50 preprocessing. The detailed high-resolution spatial information contained in the VGG16 CNN features and the global information extracted from the U-Net coding layer can ensure the accuracy of rock classification. Inspired by [6], the hybrid dilated convolution is also added to the encoding part of the network. Hybrid dilated convolution enables the detection of larger targets by expanding receptive fields, a feature that helps to detect larger rocks in images of Moon.

Figure 4 shows the overall structure of the MoonRockNet. The VGG16 [7] and U-Net are combined as the coding path of the network, while added a hybrid dilation convolution layer to the original U-Net architecture. The input of the network is a three-channel RGB image with a resolution of $512 \times 512$, and the output is a 1D array represent the rocks type predicted by the network.

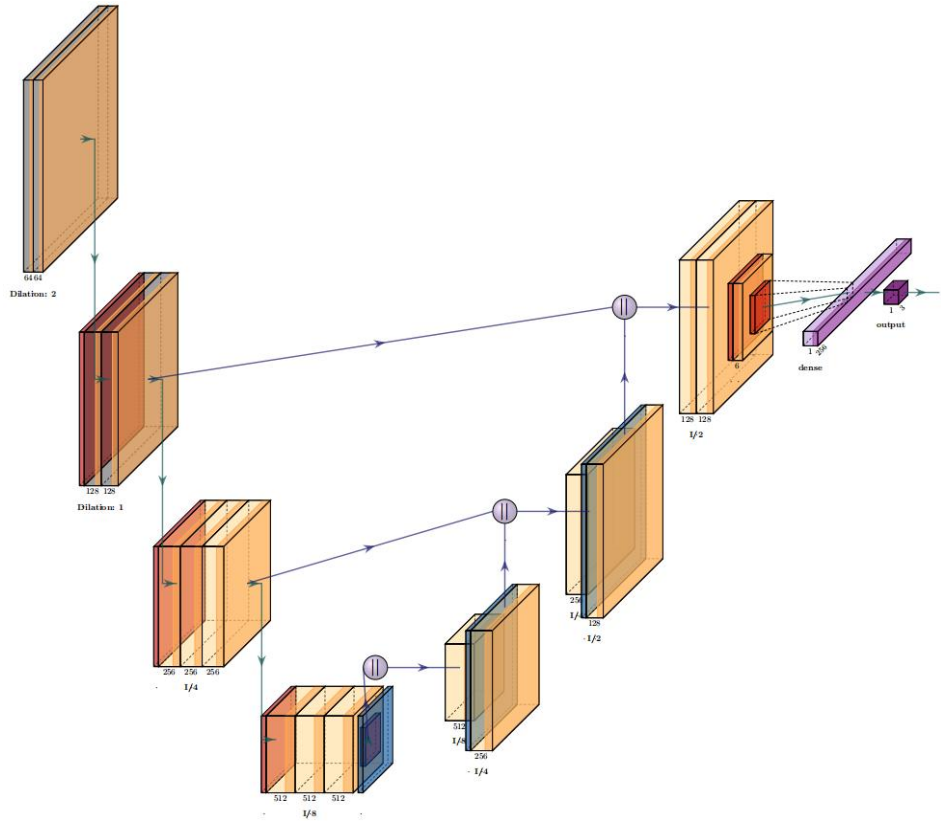

**Figure 1.** MoonRockNet structure

## 3. Experiments

*3.1 Datasets*

In the Moon surface photography, the navigation and terrain camera is mainly used by Chang'e 5. The navigation terrain camera is a binocular camera, each with a high-speed CMOS image sensor of $2048 \times 2048$ pixels. The camera has a focal length of 13.1 mm, enabling color imaging [8] of objects at a distance of 0.5m to infinity. Yellow circle in Figure 5. shows the NaTeCam.

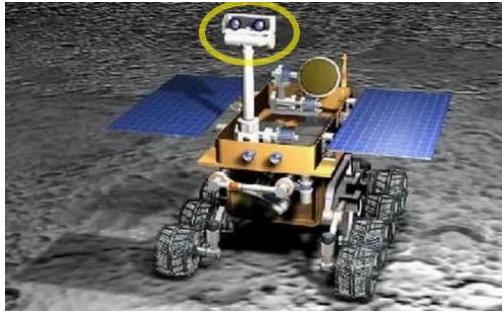

**Figure 2.** lunar rover of Chang'e 5

*3.1.1. Original data features*

The NaTeCam data for the Chang'e-5 mission uses the PDS4 standard [9], which is widely used in deep space data storage. Level 2B data product consists of data objects and data labels. Data objects store color images in Array_3D_Image format, and data labels store auxiliary information [10] of corresponding images in XML format, as shown in Table 1. We directly use python skimage package to transform all the data format to PNG.

Note that all experiments are conducted on a Google Colaboratory server with 2.30GHz Intel Xeon processor and 32GB RAM under Linux operating system, with Nvidia Tesla T4 GPU. The program codes of data processing and graph modeling are written in python version 3.6.9, tensorflow version 2.8.0, which is available for download[1][2]. In addition, Adam optimizer was used for all experiments,epoch were set to 10 and the initial learning rate was uniformly set to 0.0001, after 5 epochs the learning rate was reduced to 0.00001.

**Table 1.** Description of 2B data for the navigation terrain camera

| Category | Column2 | Column3 |
| --- | --- | --- |
| Format | Array_3D_Image | XML |
| Content | Color value of each pixel | Camera parameters and geometric position and time information |

*3.1.2. Earth rock datasets and CE5ROCK\*

- Earth rock data set

We firstly used earth rock images for model evaluation. Based on former works[11], We noticed that rock type classification is particularly challenging because of its type do not always showed on its surface,or in any visual ways. We chose 600 images of three types of rocks: igneous, metamorphic, and sedimentary. This process is done while supervised by geology major team member and under the guidance of professional geology researchers, in order to make sure the accuracy and the difference between classes are explicit. This data set is currently available for readers to download[1][3].

---

[2] 1 https://github.com/JimCui0508/Chang-e-5-rockclassification-MRNet

- CE5ROCK

Currently, 1,559 2B data for the authorized navigation and terrain camera have been acquired on May 10,2022. In order to better implement the rock classification work, the lunar surface images were specially screened to obtain 100 images with relatively well-distributed rock for subsequent model training. We also add 100 basalt and breccia rock images from National Aeronautics and Space Administration (NASA) website. We regret to inform readers that because of the unauthorization, we can not disclose a particular section of the CE5ROCK data set, if any reader is interested in this data set, please email the author for details.

*3.2 Evaluation Metrics*

In this study, we used several common evaluation metrics to measure the performance of the experimental results, Many of these metrics have been derived from the resulting confusion matrix, where TP is the prediction classified as positive has been proved to be true, FN is the prediction classified as negative has been proved to be false. The prediction classified as positive has been proved to be falsely referred to as FP, and TN is the prediction classified as negative has been proved to be true[12], the metrics mentioned as follows are shown in weighted average.

$$Fscore = \frac{2 \cdot Precision \cdot Recall}{Precision + Recall} \quad (1)$$

The area under curve score (AUC) is another popular metric in image classification. It uses receiver operating characteristics (ROC) to evaluate different thresholds to convert continuous data to discrete data for classification. It is a measure of how easily a model can distinguish between different classes[12].

*3.2.1 Loss Function*

In this paragraph we showed the loss function we used in this study for model training. We used cross entropy loss as the only loss function in our deep learning model.

$$L_{CE} = -\sum_{i=1}^{n} t_i \log(p_i) \quad (2)$$

Here, $t_i$ denotes to label, $p_i$ denotes the probability for the $ith$ class, and $n$ denotes the number of classes[12].

3.3 Experimental Results and Analysis

*3.3.1 Efficiency evaluation of the model*

We first introduced CE5ROCK data set to evaluate generalization performance of MRNet, and compared to other common models. The results are shown in Table 2. and Figure 8.

---

[3] 1 https://github.com/JimCui0508/Chang-e-5-rockclassification-MRNet

**Table 2.** general paramenters and evaluation results of each model

| | Total parameters | Time per epoch | Accuracy | Precision | Recall | Fscore |
|---|---|---|---|---|---|---|
| AlexNet | 28040483 | 5s | 0.47 | 0.30 | 0.47 | 0.36 |
| MobileNet | **3242883** | 9s | 0.20 | 0.14 | 0.37 | 0.20 |
| VGG16 | 33537363 | 26s | 0.56 | 0.57 | 0.56 | 0.55 |
| MRNet | 41366179 | 34s | **0.58** | **0.59** | **0.58** | **0.57** |

According to the final evaluation results, we can easily conclude that MRNet has the best performance in earth data set classification task, and it proves its high generalization ability.

The table shows that MobileNet had the least parameters and calculation, but also had the worst performance. The reason is MobileNet is designed for distributed devices and has remarkably reduce the complexity of the neural network, and in this case it did not work well on rock classification. Suppose we focus on on the balance between speed and accuracy. In that case, we might choose VGG16 as the best model.

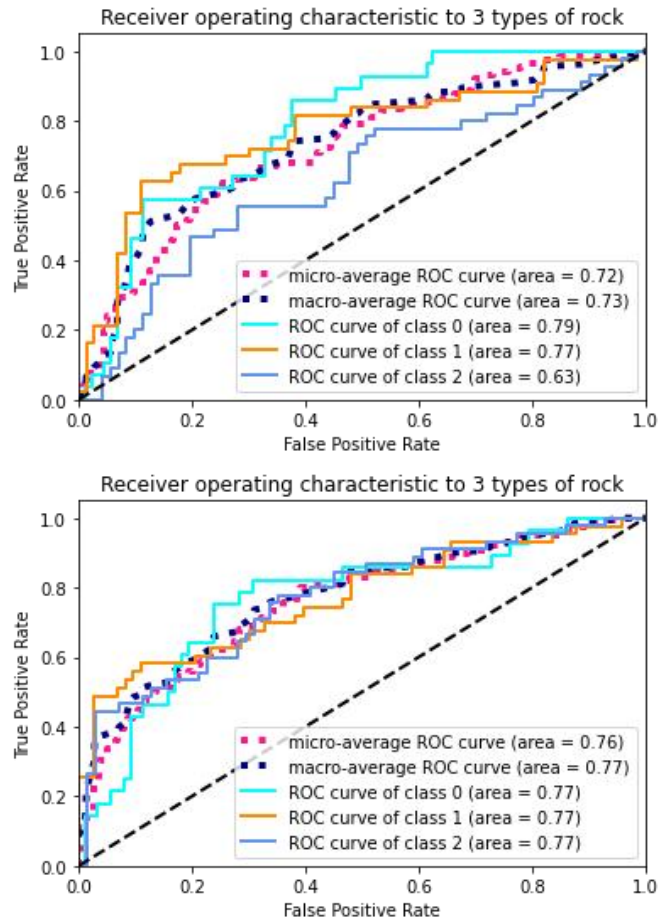

**Figure 3.** (a)ROC of VGG16, (b)ROC of MRNet

From this graph we can conclude that MRNet had a better performance than VGG16 in dealing with class imbalance, which is extremely important in defining a well-trained deep network. It had significantly improved the macro and micro average AUC. To sum up, our MRNet performed better than other common models mentioned above.

## 4. Conclusion

This paper proposed a deep learning-based lunar rock image classification network, MRNet, and evaluates its performance on different rock datasets.]In the current comparison experiments of mainstream image recognition algorithms, MRNet has the upper hand in all indicators, and the average AUC value is much higher than other networks. Compared with the combination of MobileNet and AlexNet, the accuracy of MRNet is improved by nearly 20%, which means that the network will be more suitable for engineering applications.